\documentclass[10pt]{article} 
\usepackage[preprint]{tmlr}
\usepackage{microtype}
\usepackage{graphicx}
\usepackage{subcaption}
\usepackage{booktabs}
\usepackage{array}
\usepackage{multirow}
\usepackage{fancyhdr}
\usepackage{enumitem}
\usepackage{hyperref}
\usepackage{url}
\usepackage{makecell}

\usepackage{amsmath}
\usepackage{amssymb}
\usepackage{mathtools}
\usepackage{amsthm}
\usepackage{mathrsfs} 
\usepackage{algorithm}
\usepackage{placeins}
\usepackage{needspace}
\usepackage{algpseudocode}
\usepackage{thmtools,thm-restate}

\newcommand{\Xcal}{\mathcal X}

\newcommand{\Ocal}{\mathcal O}

\newcommand{\Dcal}{\mathcal{D}}

\usepackage{hyperref}
\usepackage{url}

\title{\centering Improving Sampling Efficiency in RLVR through Adaptive Rollout and Response Reuse}

\renewcommand{\headrulewidth}{0pt}
\DeclareMathSizes{11}{11}{7}{5}

\author{\centering\normalfont\fontsize{11}{14}\selectfont
Yuheng Zhang$^{\dagger}$\thanks{Work done during the internship at Amazon.}, Wenlin Yao$^{\Diamond}$, Changlong Yu$^{\Diamond}$, Yao Liu$^{\Diamond}$ \\
Qingyu Yin$^{\Diamond}$, Bing Yin$^{\Diamond}$, Hyokun Yun$^{\Diamond}$, Lihong Li$^{\Diamond}$ \\[0.6em]
{\normalsize University of Illinois Urbana-Champaign$^{\dagger}$, Amazon$^{\Diamond}$} \\[0.3em]
{\small\texttt{yuhengz2@illinois.edu}, \texttt{\{ywenlin,changlyu,llh\}@amazon.com}}
\endgraf}

\begin{document}
\raggedbottom

\maketitle

\begin{abstract}
Group relative policy optimization (GRPO) trains large language models with verifiable rewards by normalizing rewards within response groups. When every response in a group receives the same reward, the advantages vanish and the group provides no policy gradient signal. We propose Adaptive Rollout and Response Reuse Policy Optimization (AR3PO) to obtain informative groups with fewer generations. Adaptive rollout samples responses to each prompt in stages and stops sampling once the model generates a correct response. If all sampled responses are incorrect, response reuse replaces one of them with a past correct response to the same prompt, when available, and recomputes the group advantages. Policy-recomputed reuse trains on the updated groups using probabilities recomputed under the rollout policy; advantage-only reuse uses historical responses only to compute advantages. Across four mathematical benchmarks with 7B and 8B models, AR3PO consistently improves over GRPO and achieves performance comparable to DAPO while requiring up to $4.2\times$ fewer generated responses. Experiments with a 32B model also show comparable performance to DAPO at a similar training stage with substantially fewer generated responses.
\end{abstract}

\section{Introduction}
Reinforcement learning with verifiable rewards (RLVR) has become a standard approach to improving the reasoning capabilities of large language models (LLMs)~\citep{jaech2024openai,team2025kimi,guo2025deepseek}. A verifier assigns outcome rewards to generated responses, such as a binary reward for a correct mathematical answer. Group relative policy optimization (GRPO)~\citep{shao2024deepseekmath} uses these rewards to train the model: it generates a group of responses for each prompt and normalizes their rewards within the group to compute advantages.

When all responses in a group receive the same reward, their advantages vanish and the group provides no policy gradient signal. DAPO~\citep{yu2025dapo} addresses this issue by oversampling prompts and responses and selecting groups with nonzero reward variance. This increases the number of informative groups at the cost of substantially more response generation. We ask:
\begin{center}
\it How can we obtain informative response groups with fewer generations?
\end{center}

Our approach builds on two observations. First, the model can answer some prompts correctly in fewer attempts than others, yet GRPO samples the same number of responses for every prompt. Sampling responses in stages allows us to use fewer samples for easy prompts while giving difficult prompts more attempts. Second, a correct response from an earlier training step can provide a learning signal when all current responses are incorrect. We save these past correct responses for reuse in such cases.

We propose \emph{Adaptive Rollout and Response Reuse Policy Optimization} (AR3PO), which combines these observations into a single training procedure. \emph{Adaptive rollout} samples responses to each prompt in stages and stops sampling once the model generates a correct response or reaches the maximum number of stages. For each prompt, responses from all stages form a single group for training. If all responses in that group are incorrect, \emph{response reuse} replaces one of them with a past correct response to the same prompt, when available, and recomputes the group advantages.

Because historical responses come from earlier policies, AR3PO provides two strategies for their use in policy optimization. \emph{Policy-recomputed reuse} recomputes the historical response's token probabilities under the current rollout policy and uses them in the denominator of its importance ratio, then trains on the entire updated group. \emph{Advantage-only reuse} uses the historical correct reward to give the current incorrect responses negative advantages, then trains only on those current responses. Together with adaptive rollout, these strategies make use of both new generations and previously discovered correct answers.

\begin{figure}[t]
    \centering
    \includegraphics[width=0.62\columnwidth]{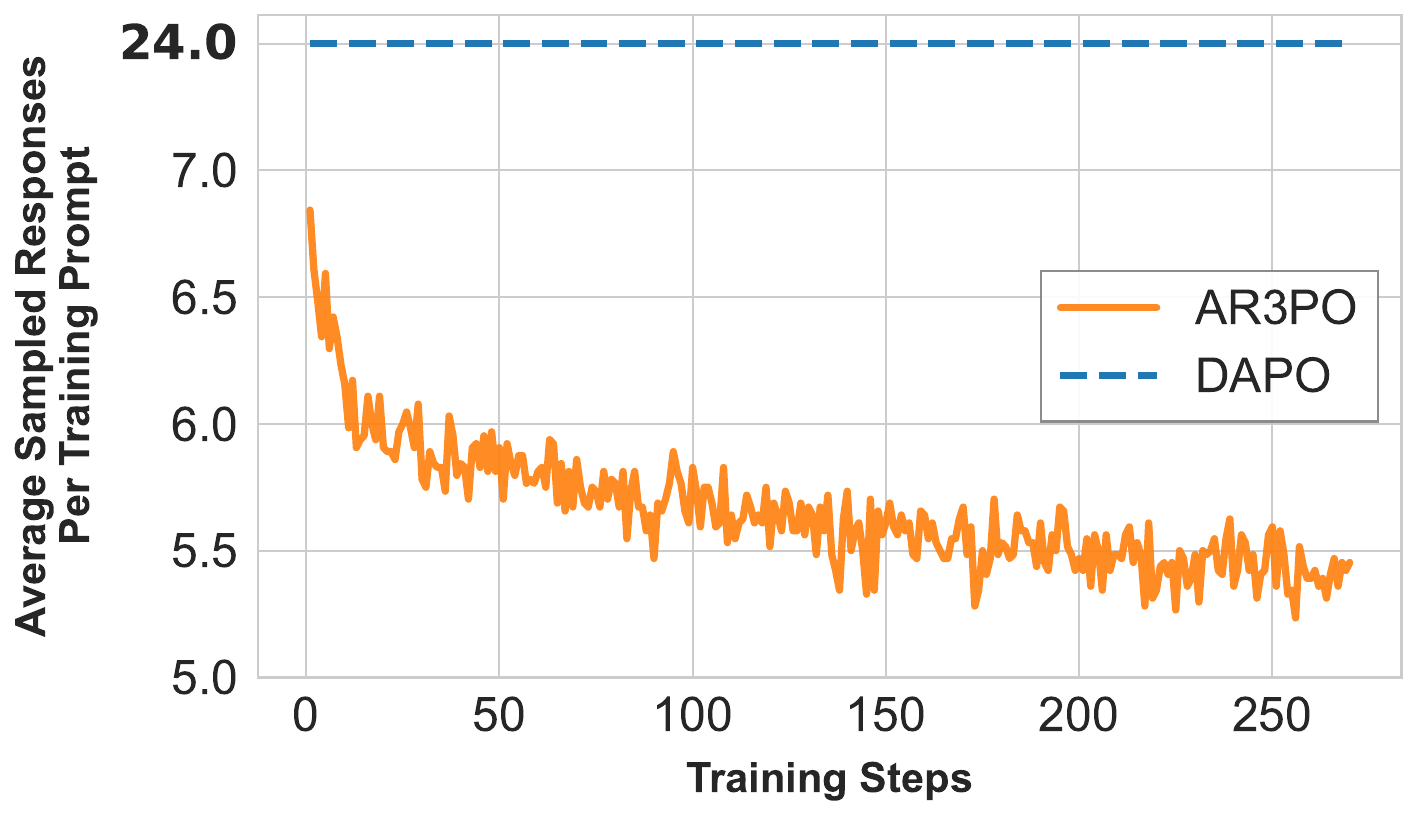}
    \caption{Average number of sampled responses per training prompt. AR3PO uses fewer responses as training progresses, reaching an average of $5.7$ and reducing the sampled response count by $4.2\times$ relative to DAPO.}
    \label{fig:algorithm}
\end{figure}

We evaluate AR3PO on four mathematical reasoning benchmarks using Qwen2.5-7B~\citep{team2024qwen2} and Llama-3.1-8B-Instruct~\citep{dubey2024llama}. AR3PO consistently improves over GRPO and achieves performance comparable to DAPO, while reducing the sampled response count by up to $4.2\times$ (Figure~\ref{fig:algorithm}). Experiments with Qwen2.5-32B show comparable performance to DAPO at a similar training stage with substantially fewer generated responses. Our analysis further shows that adaptive rollout samples fewer responses to easy prompts and that response reuse enables learning from incorrect responses by combining them with past correct responses to the same prompt.

\section{Preliminaries}\label{sec:prelim}
\paragraph{Notations.} We denote a prompt by $x\in\Xcal$ and its reference answer by $a$. An LLM is characterized by a policy $\pi_{\theta}:\Xcal \rightarrow \Delta(\Ocal)$ that maps a prompt to a distribution over the output space $\Ocal$.

\subsection{Group Relative Policy Optimization (GRPO)}
\enlargethispage{\baselineskip}
At each step, a batch of question–answer pairs $(x,a)$ is sampled from the dataset $\Dcal$, and the current policy $\pi_{\theta}$ generates a group of responses $\{o_i\}_{i=1}^G$ for question $x$.\footnote{In this paper, prompts correspond to mathematical questions fed to the LLM, and we use the terms prompt and question interchangeably.} For each response $o_i$, a math verifier provides a binary reward $R_i$ according to the answer $a$, and the advantage $A_i$ is computed by normalizing rewards within the group:
\begin{align}\label{eq:calc_adv}
A_{i}=\frac{R_i-\mathrm{mean}(\{R_i\}_{i=1}^G)}{\mathrm{std}(\{R_i\}_{i=1}^G)}.
\end{align}
When all rewards are identical, we set $A_i=0$. The GRPO objective is
\begin{equation*}
\mathcal{J}_{\textrm{GRPO}}(\theta')
= \mathbb{E}_{(x,a)\sim\mathcal{D},\,\{o_i\}_{i=1}^G\sim\pi_\theta(\cdot\mid x)}
\Bigg[
\frac{1}{\sum_{i=1}^G |o_i|}
\sum_{i=1}^G \sum_{t=1}^{|o_i|}
\min \Big(
r_{i,t}(\theta')A_i,\;
\mathrm{clip}_{[1-\epsilon^{-},\,1+\epsilon^{+}]}
\!\big(r_{i,t}(\theta')\big)A_i
\Big)
\Bigg],
\end{equation*}
\noindent where $r_{i,t}(\theta')=\frac{\pi_{\theta'}(o_{i,t} \mid x, o_{i,<t})}{\pi_{\theta}(o_{i,t} \mid x, o_{i,<t})}$ is the importance ratio. During each policy update, $\theta$ denotes the fixed rollout parameters, while $\theta'$ denotes the parameters being optimized. We use the token-level objective without a KL penalty, following~\citet{yu2025dapo}. GRPO computes advantages directly from group rewards, without a separate value network as in PPO~\citep{schulman2017proximal}. For all-correct or all-incorrect groups, the zero advantages yield no policy gradient signal.

\subsection{Dynamic Sampling Policy Optimization (DAPO)}
DAPO~\citep{yu2025dapo} uses dynamic sampling to collect groups with nonzero reward variance. It oversamples prompts and responses and selects groups containing both correct and incorrect responses. Its objective is
\begin{gather*}
\mathcal{J}_{\textrm{DAPO}}(\theta')
= \mathbb{E}_{(x,a)\sim\mathcal{D},\,\{o_i\}_{i=1}^G\sim\pi_\theta(\cdot\mid x)}
\Bigg[
\frac{1}{\sum_{i=1}^G |o_i|}
\sum_{i=1}^G \sum_{t=1}^{|o_i|}
\min \Big(
r_{i,t}(\theta')A_i,\;
\mathrm{clip}_{[1-\epsilon^{-},\,1+\epsilon^{+}]}
\!\big(r_{i,t}(\theta')\big)A_i
\Big)
\Bigg]
\\
\text{s.t.}\quad 0 < \sum_{i=1}^G R_i < G.
\end{gather*}
The sampling constraint ensures that each selected group provides nonzero advantages for the policy update.

\section{Algorithm}\label{sec:alg}
We propose \emph{Adaptive Rollout and Response Reuse Policy Optimization} (AR3PO) to obtain informative response groups with fewer generations. For each prompt, \emph{adaptive rollout} samples responses in stages and stops sampling once the model generates a correct response. If all sampled responses are incorrect after the final stage, \emph{response reuse} replaces one of them with a past correct response to the same prompt, when available. The resulting groups are then used for advantage computation and policy optimization.

\subsection{Adaptive Rollout}
At each training step, we initialize an active prompt pool $\mathcal{U}$ with the sampled prompt batch. At each of at most $S$ stages, the rollout policy $\pi_\theta$ generates $k$ responses for every $x\in\mathcal{U}$. We add these responses and their rewards to the response group $\mathcal{O}(x)$, then remove prompts with at least one correct response from $\mathcal{U}$. Generation ends when $\mathcal{U}$ is empty or all $S$ stages have been completed. All responses collected across stages remain in $\mathcal{O}(x)$ for training, including those from prompts that exit early. We write $G=|\mathcal{O}(x)|$ for the total number of responses collected for a given prompt.

Adaptive rollout uses fewer samples for easy prompts while giving difficult prompts more attempts. Once the model generates a correct response to a prompt, no additional responses are sampled for that prompt; otherwise, sampling continues in the next stage, up to the maximum of $S$ stages.

\subsection{Response Reuse}
\label{sec:response-reuse}
When adaptive rollout generates only incorrect responses to a prompt, we look for a correct response to the same prompt from an earlier training step. We maintain a prompt-indexed buffer $\mathcal{B}(x)$ of previously generated correct responses. If $\mathcal{B}(x)$ is non-empty, we randomly select $o_c\in\mathcal{B}(x)$ and replace one incorrect response in $\mathcal{O}(x)$, keeping the group size equal to $G$. Without loss of generality, let $o_G$ be the replaced response. We recompute the normalized advantages over the updated group $\{o_1,\ldots,o_{G-1},o_c\}$ using Eq.~\ref{eq:calc_adv}, denoting them by $\widetilde{A}_i$ and $\widetilde{A}_c$.

\begin{algorithm}[!t]
\fontsize{10}{12}\selectfont
\algtext*{EndIf}
\algtext*{EndFor}
\caption{Adaptive Rollout and Response Reuse Policy Optimization (AR3PO)}
\label{alg:a1}
\begin{algorithmic}
\State Initialize policy $\pi_\theta$, dataset $\mathcal{D}$, and buffers $\mathcal{B}(x)\leftarrow\emptyset$
\For{step $=1,\ldots,T$}
    \State Sample a prompt batch $\mathcal{D}_b$ from $\mathcal{D}$; set $\mathcal{U}\leftarrow\mathcal{D}_b$
    \State Initialize response groups $\mathcal{O}(x)\leftarrow\emptyset$ for each $x\in\mathcal{D}_b$
    \For{$s=1,\ldots,S$}
        \State \textbf{if} $\mathcal{U}=\emptyset$ \textbf{then break}
        \State Generate $k$ responses from $\pi_\theta(\cdot\mid x)$ for each $x\in\mathcal{U}$
        \State Verify the responses and add them with their binary rewards to $\mathcal{O}(x)$
        \State Remove prompts with at least one correct response from $\mathcal{U}$
    \EndFor
    \For{each $x\in\mathcal{U}$ with $\mathcal{B}(x)\ne\emptyset$}
        \State Replace one incorrect response in $\mathcal{O}(x)$ with $o_c\sim\mathcal{B}(x)$
    \EndFor
    \State Compute group advantages using Eq.~\ref{eq:calc_adv}
    \For{each $x\in\mathcal{D}_b$}
        \If{no response was reused for $x$}
            \State Use the GRPO objective with $\mathcal{O}(x)$
        \ElsIf{using Option I}
            \State Recompute token probabilities of $o_c$ under $\pi_\theta$; use Eq.~\ref{eq:reuse-i}
        \Else
            \State Exclude $o_c$ from the policy loss and its token normalization; use Eq.~\ref{eq:reuse-ii}
        \EndIf
    \EndFor
    \State Initialize $\theta'\leftarrow\theta$ and optimize the selected objectives over the batch
    \State Set $\theta\leftarrow\theta'$
    \State Add newly generated correct responses to $\mathcal{B}(x)$ for each $x\in\mathcal{D}_b$
\EndFor
\end{algorithmic}
\end{algorithm}

The remaining responses $\{o_i\}_{i=1}^{G-1}$ were sampled from the rollout policy $\pi_\theta$, whereas $o_c$ was generated by an earlier behavior policy $\pi_{\bar\theta}$. Direct importance-weighted replay would use the ratio
\mbox{$r_{c,t}^{\mathrm{direct}}(\theta')=\pi_{\theta'}(o_{c,t}\mid x,o_{c,<t})/\pi_{\bar\theta}(o_{c,t}\mid x,o_{c,<t})$}.
As the policy changes, using probabilities from the earlier policy in the denominator can lead to extreme importance ratios and unstable updates. We consider two strategies to stabilize reuse, both using the clipped objective $\ell(r,A)=\min(rA,\mathrm{clip}_{[1-\epsilon^{-},1+\epsilon^{+}]}(r)A)$.

\paragraph{Option I: Policy-recomputed reuse.}
We recompute the reused response's token probabilities under the fixed rollout policy $\pi_\theta$, giving
\begin{align*}
r_{c,t}^{\mathrm{I}}(\theta')
=
\frac{
\pi_{\theta'}(o_{c,t} \mid x,o_{c,<t})
}{
\pi_{\theta}(o_{c,t} \mid x,o_{c,<t})
}.
\end{align*}
At the start of the update, $\theta'=\theta$, so this ratio equals one. The resulting objective trains on both the current responses and the historical correct response:
\begin{align}
\mathcal{J}_{\mathrm{I}}(\theta')
=
\mathbb{E}\!\left[
\frac{1}{Z}
\left(
\sum_{i=1}^{G-1}\sum_{t=1}^{|o_i|}
\ell(r_{i,t}(\theta'),\widetilde{A}_i)
+
\sum_{t=1}^{|o_c|}
\ell(r_{c,t}^{\mathrm{I}}(\theta'),\widetilde{A}_c)
\right)
\right],
\label{eq:reuse-i}
\end{align}
where $Z=\sum_{i=1}^{G-1}|o_i|+|o_c|$ is the total number of tokens in the updated group.

The reused response $o_c$ was generated by the earlier behavior policy $\pi_{\bar\theta}$, whereas the denominator of its importance ratio uses token probabilities under the fixed rollout policy $\pi_\theta$. This introduces bias relative to direct importance-weighted replay, but mitigates extreme ratios caused by a large mismatch between the earlier behavior policy and the current rollout policy.

\Needspace{6\baselineskip}
\paragraph{Option II: Advantage-only reuse.}
We use the historical correct response to compute group advantages, while only the current responses enter the policy loss. Its positive reward raises the group mean, giving the current incorrect responses negative advantages and a learning signal to reduce their probabilities. The objective is
\begin{align}
\mathcal{J}_{\mathrm{II}}(\theta')
=
\mathbb{E}\!\left[
\frac{1}{Z_{\mathrm{on}}}
\sum_{i=1}^{G-1}\sum_{t=1}^{|o_i|}
\ell(r_{i,t}(\theta'),\widetilde{A}_i)
\right],
\label{eq:reuse-ii}
\end{align}
where $Z_{\mathrm{on}}=\sum_{i=1}^{G-1}|o_i|$ counts only the tokens in the current responses.

When no response is reused for a prompt, we apply the GRPO objective to its sampled responses. At the end of each training step, newly generated correct responses are added to $\mathcal{B}(x)$ for future reuse.

Algorithm~\ref{alg:a1} summarizes the complete procedure.

\FloatBarrier
\section{Experiments}

\begin{table*}[ht]
\centering
\begin{tabular}{l l c c c c c c c}
\toprule
Model & Method & 
\makecell{Math\\500} & 
\makecell{Minerva\\Math} & 
\makecell{Olympiad\\Bench} & 
\makecell{AIME\\24} & 
Average & 
\makecell{Sampled\\Responses} &
\makecell{Response Count\\Reduction\\(vs. DAPO)} \\
\midrule
\multirow{3}{*}{Qwen}
  & GRPO & 77.5 & 37.4 & 38.8 & 15.2 & 42.2 & 512 $\times$ 8.0 & 3.0 \\
  & DAPO & 77.2 & 36.4 & 41.1 & 16.9 & 42.9 & 1536 $\times$ 8.0 & 1.0 \\
  & AR3PO & 78.8 & 36.0 & 39.6 & 18.0 & \textbf{43.1} & 512 $\times$ 5.7 & \textbf{4.2} \\
\midrule
\multirow{3}{*}{Llama}
  & GRPO & 52.6 & 26.6 & 19.9 & 6.7 & 26.5 & 512 $\times$ 8.0 & 3.0 \\
  & DAPO & 53.6 & 28.1 & 20.3 & 9.2 & 27.8 & 1536 $\times$ 8.0 & 1.0 \\
  & AR3PO & 53.7 & 26.6 & 21.2 & 9.5 & \textbf{27.8} & 512 $\times$ 6.7 & \textbf{3.6} \\
\bottomrule
\end{tabular}
\caption{Comparison of AR3PO with baselines on four mathematical benchmarks. 
The penultimate column reports the average number of sampled responses per training step 
(computed as generation batch size $\times$ number of rollouts), 
and the last column reports the corresponding reduction factor relative to DAPO.}
\label{tab:main_results}
\end{table*}

\begin{figure*}[ht]
    \centering
    \begin{subfigure}{0.48\linewidth}
        \centering
        \includegraphics[width=\linewidth]{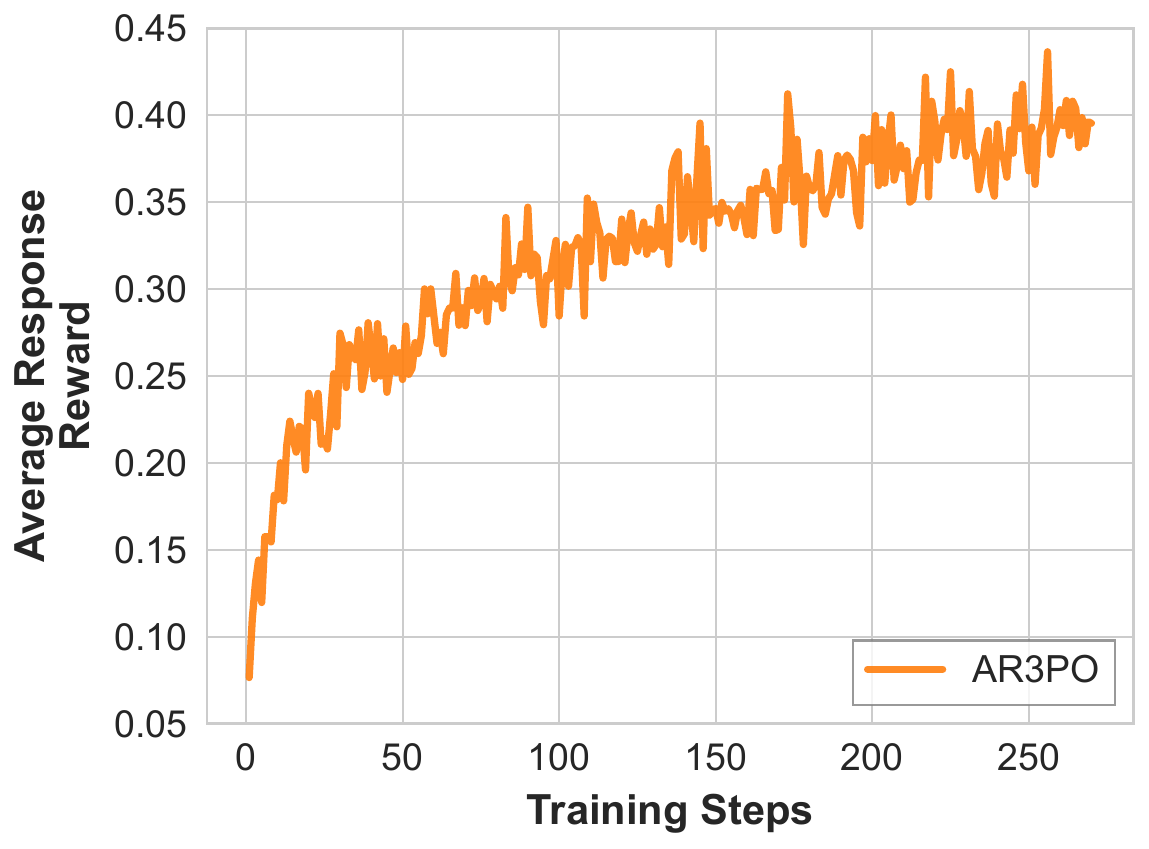}
        \caption{Reward score.}
        \label{subfig:reward}
    \end{subfigure}
    \hfill
    \begin{subfigure}{0.48\linewidth}
        \centering
        \includegraphics[width=\linewidth]{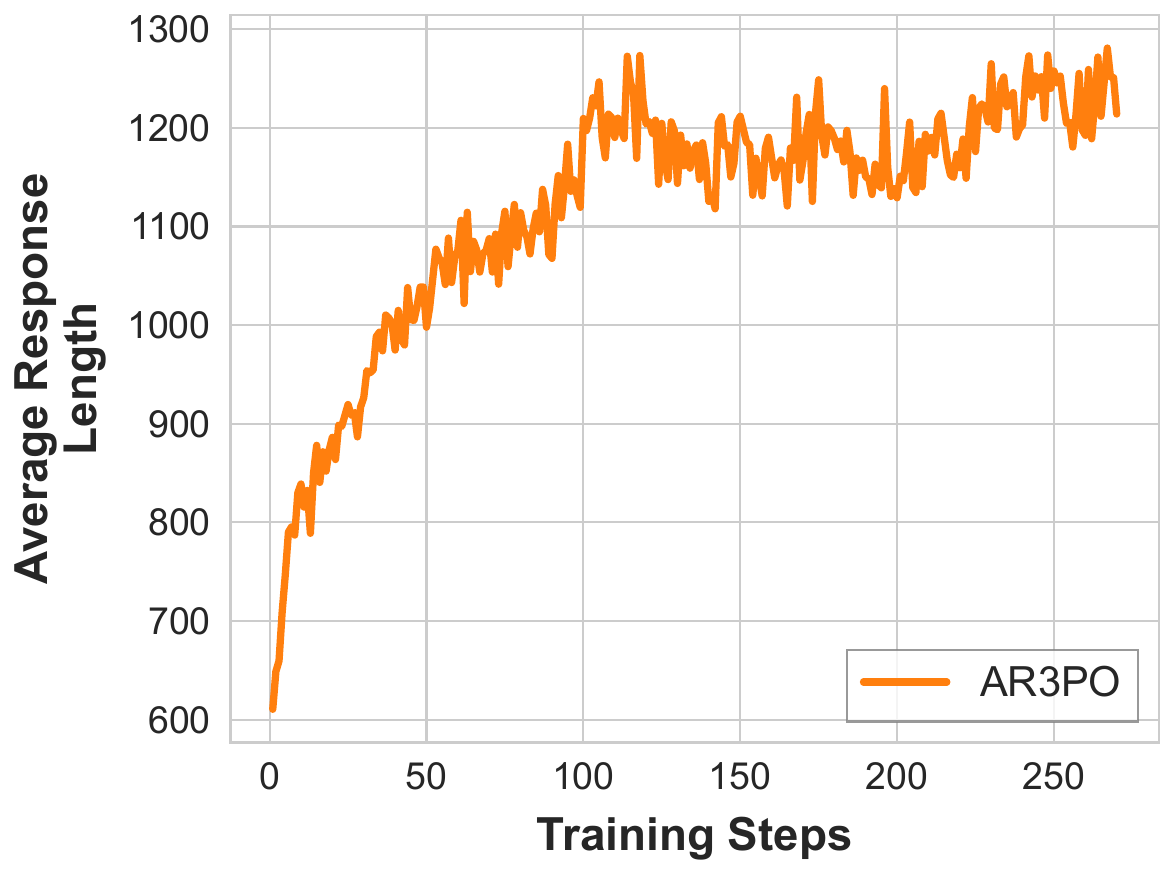}
        \caption{Response length.}
        \label{subfig:length}
    \end{subfigure}
    
    \vspace{0.4em} 
    
    \begin{subfigure}{0.48\linewidth}
        \centering
        \includegraphics[width=\linewidth]{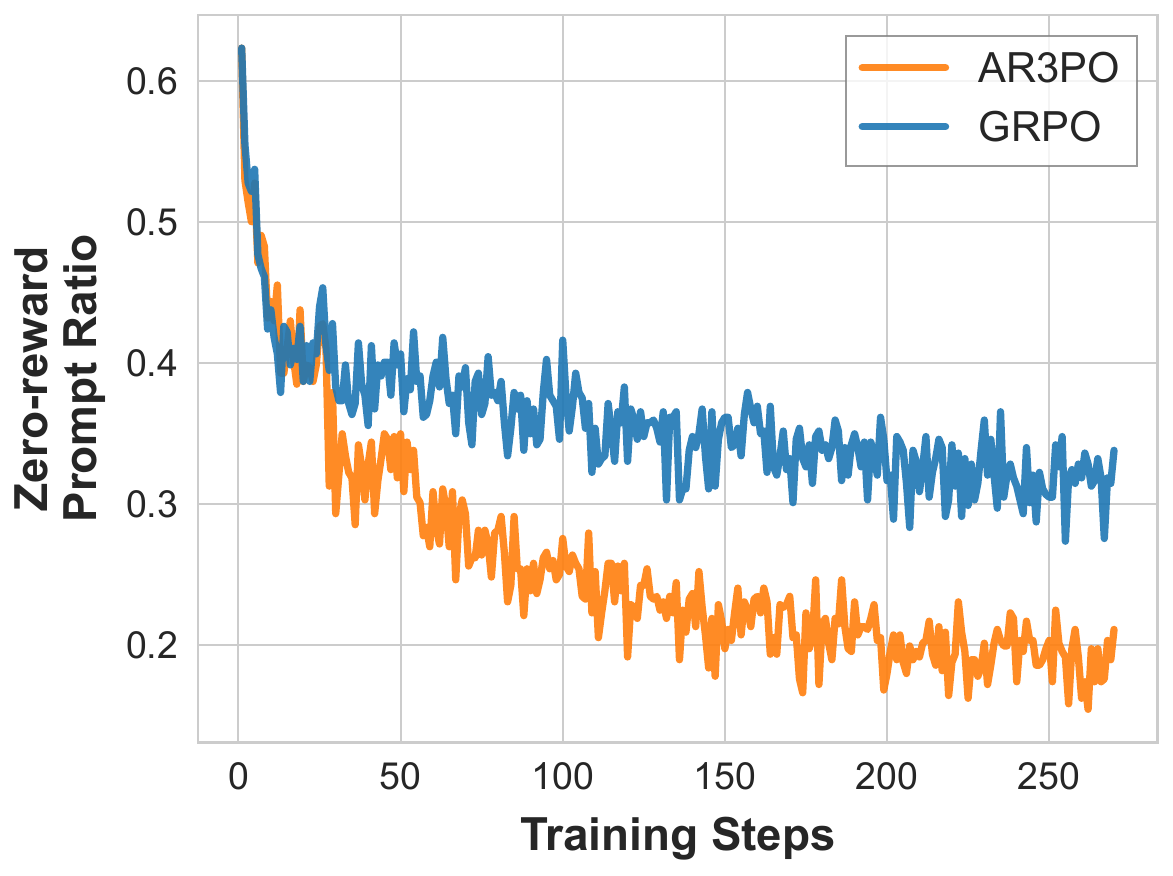}
        \caption{Zero-reward prompt ratio.}
        \label{subfig:ratio}
    \end{subfigure}
    \hfill
    \begin{subfigure}{0.48\linewidth}
        \centering
        \includegraphics[width=\linewidth]{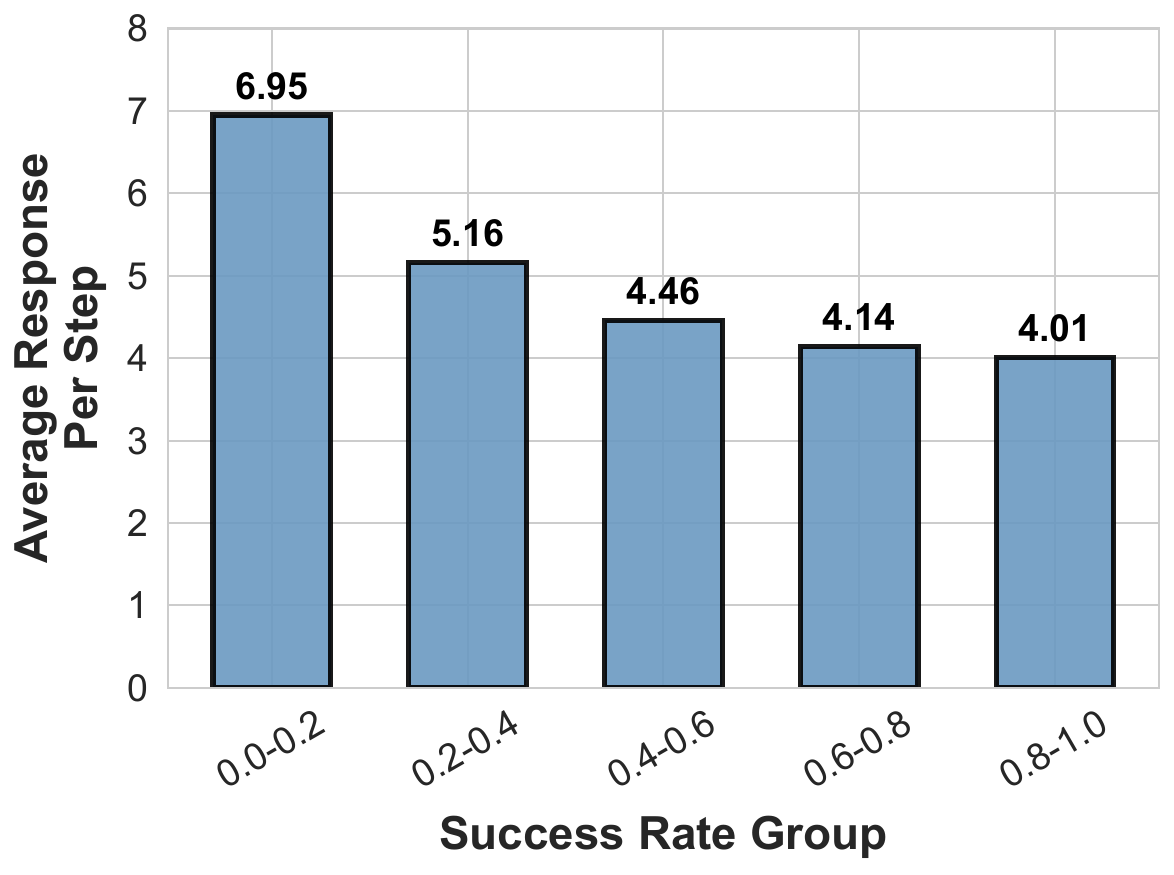}
        \caption{Response allocation.}
        \label{subfig:dist}
    \end{subfigure}
    
    \caption{(a) and (b) show the metric curves of average reward and response length for AR3PO. (c) shows the ratio of prompts without any correct response in the group. (d) presents the average sampled responses per step for prompts grouped by cumulative success rate.}
    \label{fig:metrics}
\end{figure*}

\subsection{Experimental Setup}\label{sec:exp_set}
\paragraph{Datasets and Models.} 
We focus on mathematical reasoning tasks and adopt the DAPO-Math dataset~\citep{yu2025dapo} as our training set. The original dataset contains 17K prompts, each associated with an integer answer. After removing non-English prompts, 14K prompts remain. We evaluate model performance on four representative mathematical benchmarks: Math500~\citep{hendrycks2021measuring}, Minerva Math~\citep{lewkowycz2022solving}, Olympiad Bench~\citep{he2024olympiadbench}, and AIME 2024\footnote{\url{https://huggingface.co/datasets/Maxwell-Jia/AIME_2024}}. 
OlympiadBench includes multimodal questions in both mathematics and physics; we restrict our evaluation to the text-only mathematics subset. 
For Math500, Minerva Math, and OlympiadBench, we report \texttt{avg@8}, as these datasets contain hundreds of problems. 
For AIME 2024, which contains only 30 problems, we report \texttt{avg@64}. We use Math-Verify\footnote{\url{https://github.com/huggingface/Math-Verify}} as the verifier to check the correctness of the output answer. To better assess the generality of our algorithms, we conduct experiments on models from different families: Qwen2.5-7B~\citep{team2024qwen2} and Llama-3.1-8B-Instruct~\citep{dubey2024llama}.

\paragraph{Implementation Details.}
We compare AR3PO against GRPO~\citep{shao2024deepseekmath} and DAPO~\citep{yu2025dapo}, with all methods implemented in veRL~\citep{sheng2025hybridflow}. We use a training prompt batch size of 512 and a mini-batch size of 128 for gradient updates. DAPO uses a data prompt batch size of 1536 for dynamic sampling, following its official implementation. GRPO and DAPO generate 8 responses per prompt, while AR3PO uses $S=2$ and $k=4$, generating at most 8 responses per prompt.

For both models, the maximum prompt length is 1024. For Qwen2.5-7B, we use a maximum response length of 3072 and a learning rate of $1\times 10^{-6}$. For Llama-3.1-8B-Instruct, we use a maximum response length of 2048 and a smaller learning rate of $1\times 10^{-7}$, as the model has already undergone RLHF post-training. We use no learning rate warmup and follow the default veRL settings for all other configurations. Shared training hyperparameters are held fixed across methods.

\subsection{Main Results}
We use \emph{advantage-only reuse} for the 7B and 8B experiments. Table~\ref{tab:main_results} reports results on four mathematical benchmarks. AR3PO achieves average scores of $43.1$ and $27.8$ on Qwen2.5-7B and Llama-3.1-8B-Instruct, respectively, improving over GRPO and matching or exceeding DAPO's average scores. It uses $4.2\times$ and $3.6\times$ fewer generated responses per training step than DAPO. In the Qwen2.5-7B setup, average generation time per step is approximately 200s for AR3PO, 250s for GRPO, and 500s for DAPO.

Adaptive rollout reduces generation cost by stopping early on easy prompts while giving difficult prompts more attempts. Response reuse complements this allocation by allowing difficult prompts to provide informative training signals even when all responses generated in the current step are incorrect, provided a historical correct response is available.

\subsection{Analysis of Response Reuse Strategies}
\begin{table*}[ht]
\centering
\begin{tabular}{l c c c c c}
\toprule
Method &
\makecell{Math\\500} &
\makecell{Minerva\\Math} &
\makecell{Olympiad\\Bench} &
\makecell{AIME\\24} &
Average \\
\midrule
AR3PO w/ direct rollout replay & 77.3 & 35.7 & 39.1 & 15.0 & 41.8 \\
AR3PO w/ policy-recomputed reuse             & 76.9 & 35.8 & 39.3 & 19.3 & 42.8 \\
AR3PO w/ advantage-only reuse             & 78.8 & 36.0 & 39.6 & 18.0 & \textbf{43.1} \\
\bottomrule
\end{tabular}
\caption{Comparison of response reuse strategies with Qwen2.5-7B. Direct rollout replay uses behavior-policy probabilities in the reused response's importance ratio. The two AR3PO strategies are defined in Section~\ref{sec:response-reuse}.}
\label{tab:comparison}
\end{table*}

Table~\ref{tab:comparison} compares the two AR3PO reuse strategies with direct rollout replay under the same adaptive rollout procedure. Policy-recomputed reuse raises the average score from $41.8$ to $42.8$, while advantage-only reuse achieves the highest average score of $43.1$. Policy-recomputed reuse achieves the highest AIME 2024 score of $19.3$. These results show that both strategies improve over direct replay; advantage-only reuse attains the strongest overall performance, and policy-recomputed reuse performs particularly well on AIME.

\begin{figure}[t]  
    \centering
    \includegraphics[width=0.62\columnwidth]{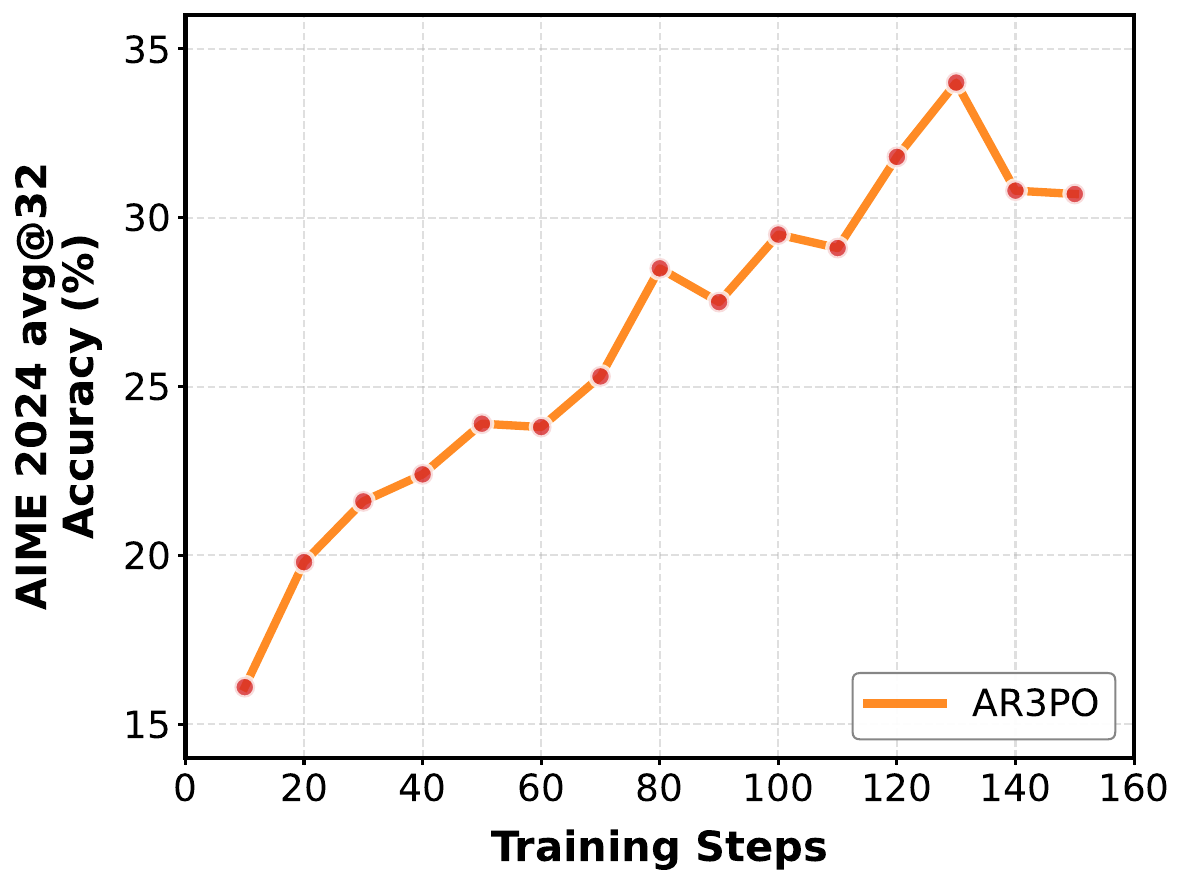}
    \caption{AIME 2024 accuracy of AR3PO with the Qwen2.5-32B base model. Our method achieves $34.0\%$ accuracy at step 130, comparable to the performance reported by DAPO around 130 training steps (see Figure~1 in~\citet{yu2025dapo}). Note that their figure reports gradient update steps, with 16 updates per step; therefore, our step 130 roughly corresponds to their reported step 2000.}
    \label{fig:aime}
\end{figure}

\subsection{Training Dynamics}

Figure~\ref{subfig:reward} reports the average reward of AR3PO's generated responses at each step. The reward increases steadily throughout training, indicating stable optimization. As training progresses and more responses are allocated to difficult prompts, the reward growth becomes slower than in the initial phase. Figure~\ref{subfig:length} shows the average response length during training. We observe that the length increases rapidly at the beginning and then fluctuates around 1200. This is consistent with prior findings~\citep{guo2025deepseek,yu2025dapo}, which report that response length does not always increase monotonically and may even decrease during training. Because adaptive rollout changes the sampled response distribution, these reward and length statistics are not directly comparable with those of GRPO and DAPO.

Figure~\ref{subfig:ratio} presents the ratio of prompts without any correct response for AR3PO and GRPO. 
The curves nearly overlap during the first 30 steps, which corresponds to the first training epoch. 
This is expected because most prompts have not yet been revisited, so response reuse is rarely available. 
Afterward, as previously collected correct responses become available, response reuse is invoked for about $10\%$ of prompts in later steps, and AR3PO reduces the ratio from about $0.3$ to below $0.2$, while GRPO remains around $0.3$. 
This gap indicates that response reuse is actively triggered after prompts begin to repeat, providing additional training signals for difficult prompts.

For prompts grouped by cumulative success rate, Figure~\ref{subfig:dist} reports the average number of generated responses. For prompts with a cumulative success rate from 0.0 to 0.2, AR3PO samples an average of 6.95 responses per step, compared with about 4 for prompts with a success rate from 0.8 to 1.0. These results show that adaptive rollout samples fewer responses to easy prompts while giving difficult prompts more attempts.

\subsection{Results on Larger Model}
\label{sec:32b}

To further evaluate whether AR3PO scales to larger models, we conduct experiments with the Qwen2.5-32B base model, as shown in Figure~\ref{fig:aime}. 
We use \emph{policy-recomputed reuse}, which achieves the highest AIME 2024 score in Table~\ref{tab:comparison}. 
We train the model for 150 steps. 
AR3PO achieves $34.0\%$ accuracy on AIME 2024 at step 130, which is comparable to the performance reported by DAPO at a similar training stage. 
Relative to the released DAPO generation configuration of 1,536 prompts and 16 rollouts per prompt,\footnote{Public DAPO configuration: \href{https://github.com/verl-project/verl/blob/4f80e465c2ec79ab9c3c30ec74b9745de61d0490/recipe/dapo/run_dapo_qwen2.5_32b.sh}{verl commit \texttt{4f80e4}, \texttt{run\_dapo\_qwen2.5\_32b.sh}}.} AR3PO uses a generation batch size of 512 and 5.3 responses per prompt on average. 
This corresponds to about a $9\times$ reduction in the number of generated responses, indicating that the sampling efficiency benefit remains substantial at the 32B scale.

\FloatBarrier
\section{Related Work}
\paragraph{RL algorithms for LLM post-training.} 
Since the success of reinforcement learning from human feedback (RLHF) in ChatGPT~\citep{ouyang2022training,achiam2023gpt}, reinforcement learning algorithms have been extensively explored for LLM post-training. 
To align LLMs with human preferences, \citet{bai2022training} employ the PPO algorithm~\citep{schulman2017proximal} to optimize a KL-regularized objective. 
\citet{rafailov2024direct} propose direct preference optimization (DPO), which directly minimizes a loss function derived from the Bradley–Terry (BT) model to capture human preferences. Building on DPO, a number of variants have been developed, including offline algorithms such as KTO~\citep{ethayarajh2024kto}, ORPO~\citep{hong2024orpo}, and SimPO~\citep{meng2024simpo}, as well as online algorithms such as iterative DPO~\citep{dong2024rlhf} and XPO~\citep{xie2024exploratory}. 
In addition, a line of work has investigated general preference alignment methods that relax the BT model assumption~\citep{azar2024general,munos2023nash,ye2024online,wu2024self,zhang2024iterative,zhang2025improving}.

RLVR trains reasoning models with verifiable outcome rewards~\citep{jaech2024openai,guo2025deepseek}. GRPO~\citep{shao2024deepseekmath} computes advantages from group-normalized rewards, and subsequent variants modify the importance ratio or advantage computation~\citep{liu2025understanding,zheng2025group}. DAPO~\citep{yu2025dapo} introduces dynamic sampling to select groups with nonzero reward variance. AR3PO improves sampling efficiency through staged generation and targeted reuse of historical correct responses.

\paragraph{Adaptive Rollout.} 
Recent works have explored adaptive rollout strategies for GRPO and related algorithms. 
\citet{liao2025enhancing} allocate rollout budgets based on the ranking of prompts by their historical average reward, giving more budget to difficult prompts to encourage correct responses. 
\citet{yao2025optimizing} estimate the required number of rollout samples to minimize gradient variance in the optimization objective. 
A concurrent work~\citep{yang2025depth} also designs a rollout allocation method based on cumulative accuracy. 
Our goal is to reduce generation cost while maintaining reasoning performance and preserving informative training signals for difficult prompts. AR3PO samples responses in stages and stops sampling for a prompt once the model generates a correct response. It makes this decision using responses from the current rollout, so easy prompts require fewer samples while difficult prompts get more attempts.

\paragraph{Rollout Replay.} 
Several concurrent works have explored rollout replay in RLVR to improve reasoning performance. 
\citet{sun2025improving} reuse recent rollout responses from a replay buffer to update the policy, while \citet{zhang2025rlep} sample previous correct responses and combine them with rollouts from the current policy during training. 
AR3PO reuses a past correct response when all responses sampled for the same prompt in the current step are incorrect. Policy-recomputed reuse includes the past response in the policy loss after recomputing its probabilities under the rollout policy. Advantage-only reuse uses its reward to compute group advantages, while only the current responses enter the policy loss.

\section{Conclusion and Future Work}
AR3PO improves the sampling efficiency of RLVR by combining adaptive rollout with response reuse. For each prompt, AR3PO samples responses in stages and stops sampling once the model generates a correct response. If all sampled responses are incorrect, it reuses a past correct response to the same prompt, when available, to compute group advantages. Policy-recomputed and advantage-only reuse provide two ways to incorporate these answers into training. Across mathematical reasoning benchmarks, AR3PO improves over GRPO and achieves performance comparable to DAPO with substantially fewer generated responses. Future work may extend these ideas to broader LLM agent settings, where sampling efficiency is increasingly important.

\bibliography{main}
\bibliographystyle{tmlr}

\end{document}